\documentclass[11pt,a4paper]{article}
\usepackage[hyperref]{emnlp-ijcnlp-2019}
\usepackage{times}
\usepackage{latexsym}
\usepackage{breakurl}

\usepackage{amsmath}
\usepackage{amssymb}
\usepackage{bm}
\usepackage{todonotes}
\usepackage{titlesec}
\usepackage{pgfplots}
\pgfplotsset{compat=newest}

\aclfinalcopy % Uncomment this line for the final submission
\title{Autoregressive Text Generation Beyond Feedback Loops}

\author{
  Florian Schmidt \\
  Department of Computer Science\\
  ETH Z{\"u}rich\\
  \texttt{florian.schmidt@inf.ethz.ch} \\
  \And
  Stephan Mandt \\
  Department of Computer Science\\
  University of California, Irvine\\
  \texttt{mandt@uci.edu} \\
  \AND
  Thomas Hofmann \\
  Department of Computer Science\\
  ETH Z{\"u}rich\\
  \texttt{thomas.hofmann@inf.ethz.ch} \\
}

\date{}

\newcommand\h{\mathbf h}

\newcommand\x{\mathbf x}
\newcommand\oo{\mathbf o}

\newcommand\E{\mathbb E}
\newcommand\w{w}

\newcommand\Ts{{1:T}}

\newcommand\A{\mathbf A}
\newcommand\X{\mathbf X}
\newcommand\Y{\mathbf Y}
\newcommand\T{\mathbf T}
\newcommand\R{\mathbb R}

\renewcommand\S{\mathbf S}

\newcommand\bbeta{{\bm\beta}}

\let\oldpm\pm
\renewcommand\pm{\ensuremath{\oldpm}}

\usetikzlibrary{positioning,arrows,shapes.geometric,calc, matrix}

\definecolor{rosso}{RGB}{220,57,18}
\definecolor{giallo}{RGB}{255,153,0}
\definecolor{blu}{RGB}{102,140,217}
\definecolor{verde}{RGB}{16,150,24}
\definecolor{viola}{RGB}{153,0,153}
\definecolor{babyblue}{RGB}{0,129,255}
\definecolor{darkgreen}{RGB}{6,148,60}

\begin{document}

\maketitle
\begin{abstract}
Autoregressive state transitions, where predictions are conditioned on past predictions, are the predominant choice for both deterministic and stochastic sequential models. However, autoregressive feedback exposes the evolution of the hidden state trajectory to potential biases from well-known train-test discrepancies. In this paper, we combine a latent state space model with a CRF observation model. We argue that such autoregressive observation models form an interesting middle ground that expresses local correlations on the word level but keeps the state evolution non-autoregressive. 
On unconditional sentence generation we show performance improvements compared to  RNN and GAN baselines while avoiding some prototypical failure modes of autoregressive models.\footnote{Code and generated sentences available at \url{https://github.com/schmiflo/crf-generation}}
\end{abstract}

\section{Introduction}
Sequential autoregressive models express predictions of observations based on past predictions.  
They are the predominant architecture for text generation in a maximum likelihood setup \cite{graves2013generating, sutskeverVL2014} and are used in machine translation \cite{bahdanauCB14,vaswaniSPUJGKP17}, summarization \cite{rushCW15}, and dialogue systems \cite{serban2016building}.

An immediate consequence of combining autoregressive modeling and maximum likelihood training is that past observations enter the loss functions as ground-truth, not predicted observations \cite{dlbook}. This discrepancy is often summarized as \emph{teacher-forcing} and the bias it implies is referred to as  \emph{exposure-bias} \cite{ranzato2015sequence, goyalLZZCB16}. 

The standard methodology to turn a sequential model into an autoregressive one is to introduce a \emph{feedback loop}, where one provides the last predicted token as a feature to the computation of the next state \cite{graves2013generating}. The ground-truth observations become effectively input features for the evolution of the hidden state trajectory at training time. Several attempts have been made to introduce robustness with respect to the model's predictions by leaving the maximum likelihood framework, either implicitly \cite{bengioVJS15, bowmanVVDJB15} or explicitly \cite{goyalLZZCB16, leblondAOL17}. Nevertheless, the same feedback mechanisms have been adopted in latent sequential models where they obfuscate the true stochasticity of transitions during training. Non-autoregressive sequence models have recently regained attention for unconditional \cite{schmidt18, ziegler19} and conditional \cite{lee18} generation.

We argue that there is an interesting intermediate regime between feedback-driven autoregressive models and completely non-autoregressive models, namely modeling temporal correlations as part of the \emph{observation model}. We propose a neural CRF observation model that leverages word-embeddings to explain local word correlations in a global sequence score. We show how training and generation can be performed efficiently. The result is an autoregressive model that keeps the hidden state evolution less affected by observation noise while generating coherent word sequences. 

\section{Related Work}
Conditional Random Fields (CRF) were originally introduced by \citet{sha03} to overcome \emph{label bias}, a shortcoming of locally normalized observation models. They have been applied and integrated into neural-network architectures \cite{ma16, huang15} in various sequence labeling tasks \cite{goldmanG17} where the observation space exhibits small cardinality (typically tens to hundreds).

The importance of global normalization for sequence generation has only lately been emphasized, most notably by \citet{wisemanR16} for conditional generation in a learning-as-search-optimization framework and by \cite{andorAWSPGPC16} for parsing.

Word-embeddings have been reported as excellent dense representations of sparse co-occurrence statistics within several learning frameworks \cite{mikolov13, pennington14}.  Using embeddings in pairwise potentials has been proposed by \citet{goldmanG17}, but they do not compute the true log-likelihood during training as we do. Similar techniques have been applied for various  message passing schemata \cite{kimDHR17, domke13}.

Local correlations such as our pairwise potentials have been used by \cite{noraset18}, yet as an auxiliary loss and not for model design.

Other approaches to tackle teacher-forcing have been proposed in an adversarial setting \cite{goyalLZZCB16}, in search based optimization \cite{leblondAOL17} and in a reinforcement learning setting \cite{rennieMMRG16}.

\section{Model}

Latent sequential models for text generation typically consist of two parts: A mechanism for generating a latent hidden state trajectory $\h=\h_\Ts$, and an observation model. The latter predicts the data $\w=w_\Ts$ given the latent states. The most simple dependency structure for such a model is that of an Hidden Markov Model, which breaks into transitions $p(\h_t|\h_{t-1})$ and observations $p(w_t|\h_t)$. In contrast, models with \emph{autoregressive transitions} factorize as
\begin{align}
	p(\w,\h)=\prod_{t=1}^T p(w_t|\h_t)p(\h_t|\h_{t-1},w_{t-1})\ .\label{eq:ar-joint}
\end{align}
The result is a next-state distribution with dependencies identical to deterministic RNN transitions $\h_t=F(\h_{t-1}, w_{t-1})$ and indeed similar neural networks can be used to parametrize a simple, e.g., Gaussian distribution \cite{fraccaroSPW2016}.

As a negative consequence, we inherit teacher-forcing. This comes with aforementioned biases and also conflicts with our notion of uncertainty in $p(\h_t|\h_{t-1},w_{t-1})$ which during training solely depends on the continuous parameters (i.e.\ a mean and a variance), but is greatly affected by the discrete sampling noise in $w_{t-1}$ at test time.

\tikzstyle{mycon}=[dash pattern=on 2pt off 2pt, gray]
\tikzstyle{boxstyle}=[minimum height=1cm, minimum width=4.0cm,rounded corners,fill=black!10,inner sep=0ex, yshift=-2mm]

\begin{figure*}[ht]
\centering
\newcommand\fillnode{\hspace{0.1mm}}
\begin{tikzpicture}[probnode/.style={inner sep=1pt,minimum size=4pt, font=\small	}]	
  \small
  \matrix (m) [matrix of nodes, row sep=2em, column sep=2.7em, nodes={probnode}]
       { $w_1$ & $w_1$ &  $w_1$&   $w_1$ \\
         $w_2$ & $w_2$ &  $w_2$&   $w_2$ \\  
         $w_3$ & $w_3$ &  $w_3$ &  $w_3$ \\  };
  
  % Hs
  \foreach \t in {1, ..., 4}{
  	\node[above = 1em of m-1-\t] (h-\t) {$\h_\t$};
  	\draw[->] (h-\t) -- (m-1-\t);
  }
  % horizontal H connection
  \foreach \t in {1, ..., 3}{
    \pgfmathtruncatemacro{\tplusone}{\t+1}
  	\draw[->] (h-\t) -- (h-\tplusone);
  }
  
    \node[boxstyle,align=center,anchor=north] at (current bounding box.south) {\scriptsize $\displaystyle p(\w)=\prod_{t=1}^T \frac{\exp\psi(w_t;\h_t)}{\sum_{w_t'}\exp\psi(w_t';\h_t)}$};

\end{tikzpicture}\hfill
\begin{tikzpicture}[probnode/.style={inner sep=1pt,minimum size=4pt, font=\small	}]	
  \small
  \matrix (m) [matrix of nodes, row sep=2.1em, column sep=2.7em, nodes={probnode}]
       { $w_1$ & $w_1$ &  $w_1$&   $w_1$ \\
         $w_2$ & $w_2$ &  $w_2$&   $w_2$ \\  
         $w_3$ & $w_3$ &  $w_3$ &  $w_3$ \\  };
  \foreach \t in {1, ..., 3}{
    \pgfmathtruncatemacro{\tplusone}{\t+1}
  	\foreach \i in {1, ..., 3}{
  		\foreach \j in {1, ..., 3}{
  	  	  \draw[mycon] (m-\i-\t) -- (m-\j-\tplusone);
  	  	}
  	 }
  }
  
  % Hs
  \foreach \t in {1, ..., 4}{
  	\node[above = 1em of m-1-\t] (h-\t) {$\h_\t$};
  	\draw[->] (h-\t) -- (m-1-\t);
  }
  % horizontal H connection
  \foreach \t in {1, ..., 3}{
    \pgfmathtruncatemacro{\tplusone}{\t+1}
  	\draw[->] (h-\t) -- (h-\tplusone);
  }
  
  \node[boxstyle,align=center,anchor=north] at (current bounding box.south) {\scriptsize $\displaystyle p(\w)=\frac{\exp\psi(\w;\h)}{\sum_{\w'}\exp\psi(\w';\h)}$};
        
\end{tikzpicture}\hfill
\begin{tikzpicture}[probnode/.style={inner sep=0pt,minimum size=0pt, font=\small}]	
  \small
  \matrix (m) [matrix of nodes, row sep=2.82em, column sep=4em, nodes={probnode}]
       { \fillnode& \fillnode & \fillnode & \fillnode \\
         \fillnode& \fillnode & \fillnode & \fillnode \\   
         \fillnode& \fillnode & \fillnode & \fillnode \\  };

  \foreach \t in {1, ..., 3}{
    \pgfmathtruncatemacro{\tplusone}{\t+1}
  	\foreach \i in {1, ..., 3}{
  		\foreach \j in {1, ..., 3}{
  	  	  \draw[mycon] (m-\i-\t) -- (m-\j-\tplusone);
  	  	}
  	 }
  }
  
   % PSEUDO Hs
  \foreach \t in {1, ..., 4}{
  	\node[above = 1em of m-1-\t] (h-\t) {\phantom{$\h_\t$}};
  }
  
  % Some colroed path
  \draw[mycon, giallo] (m-1-1) --  (m-3-2) --  (m-3-3) --  (m-3-4);
  \draw[mycon, babyblue] (m-2-1) --  (m-1-2) --  (m-2-3) --  (m-1-4);
  \draw[mycon, viola] (m-2-1) --  (m-2-2) --  (m-1-3) --  (m-2-4);
  \draw[mycon, rosso] (m-3-1) --  (m-3-2) --  (m-2-3) --  (m-2-4);

  % W labes
    \foreach \t in {1, ..., 4}{
  \pgfmathtruncatemacro{\tplusone}{\t+1}
  	\foreach \i in {1, ..., 3}{
  		\node[yshift=-2mm] (l-\i-\t) at (m-\i-\t) {$w_\i$};

  	 }
  }
  \scriptsize
  \node[right = 1em of m-1-4, rosso] (h-1) {$\h(w_{1:T}^{(1)})$};
  \node[below = 0em of h-1, babyblue, yshift=1mm] (h-2) {$\h(w_{1:T}^{(2)})$};
  \node[below = 0em of h-2, viola, yshift=1mm] (h-3) {$\h(w_{1:T}^{(3)})$};
  \node[right = 1em of m-3-4, yshift=5.5mm] (dots) {\hspace{4.5mm}$\vdots$};
  \node[right = 1em of m-3-4, giallo] (h-4) {$\h(w_{1:T}^{(|V|^T)})$};
	\small
  \node[boxstyle,align=center,anchor=north] at (current bounding box.south) {\scriptsize $\displaystyle p(\w)=\frac{\exp\psi(\h(\w))}{\sum_{w'}\exp\psi(\h(\w'))}$};

\end{tikzpicture}
\caption{Schematic comparison of differently normalized architectures. We sketch trellis diagrams for $V=\lbrace w_1,w_2,w_3\rbrace$ and $T=4$. Dashed lines indicate autoregressive dependencies in the log-likelihood computation. \textbf{Left}: Standard RNN with soft-max observations. Since the model is locally normalized, the trellis diagram does not unfold across time-steps. \textbf{Middle}: Our proposed CRF model. The potentials only span across pairs, but the normalization is global and can be computed exactly and efficiently. \textbf{Right}: An intractable globally normalized model in which fully-connected potentials $\psi(\h(\w))$ are obtained from an RNN. Computing a single $p(\w)$ would require running the RNN $|V|^T$ times. We highlight four runs for illustration.} 
\label{fig:sketch}
\end{figure*}
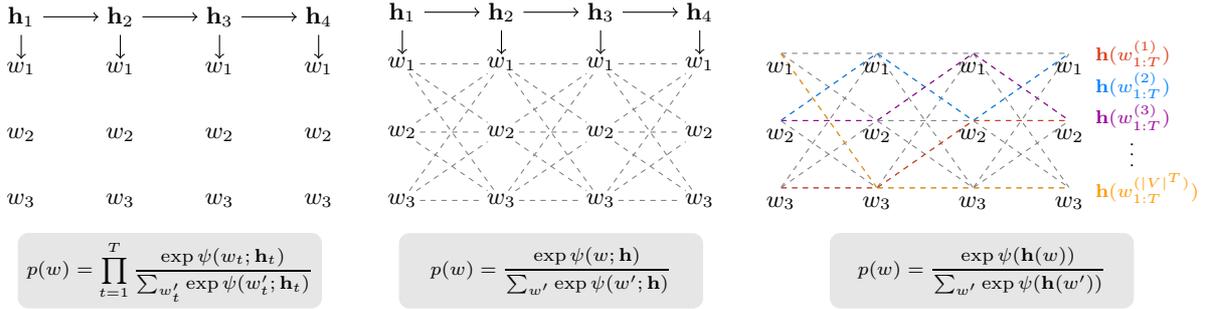

\paragraph{Autoregressive observation model}
We consider an alternative to autoregressive feedback mechanisms such as 
 \eqref{eq:ar-joint}, where predictions are directly injected into states. We write 
\begin{align}
	p(\w,\h)=p(\w|\h)\prod_{t=1}^Tp(\h_t|\h_{t-1})
\end{align}
assuming only Markovian transitions and focus on finding a powerful observation model instead. Crucially, since the state space model is not affected by previous outputs, word coherence may be lost when simply factorizing as in $p(\w|\h) = \prod_{t} p(w_t|\h_t)$, i.e.\ with independent soft-max factors $p(w_t|\h_t)\propto\exp\psi(w_t,\h_t)$ where $\psi(w_t,\h_t)=\x(w_t)^\top\h_t$. However, a natural extension can be found by  reformulating local normalization as a form of global normalization without correlations across time
\begin{align}
	p(\w|\h)&=\prod_{t=1}^T \frac{\exp\psi(w_t,\h_t)}{\sum_{w_{t}'} \exp\psi(w_t',\h_t)}\label{eq:iidsoftmax}\\
	&=\frac{\exp S(\w,\h )}{\sum_{\w'}\exp S(\w',\h)}\label{eq:global-observations}
\end{align}
where $S=\sum_{t=1}^T \psi(w_t,\h_t)$ contains no dependencies between $w_t$ and $w_{t'}$ for $t\neq t'$. As soon as we add word-correlations to $S$, we obtain a truly global observation model that cannot be expressed in the form of \eqref{eq:iidsoftmax}.

\subsection{CRF Observation Model}
\label{sec:CRF}

Equation \eqref{eq:global-observations} describes a
conditional random field (CRF) with an energy function $S$  \citep{sha03}. We consider up to pairwise interactions between consecutive words
\begin{align}
\!\!\!S(\w;\h)\!=\!\!\sum_{t=1}^T\psi(w_t;\h_t)\! +\! \psi(w_{t-1}, w_t;\h_{t-1:t})\!\!\label{eq:energy-function}
\end{align} 
The potentials $\psi$ reflect the independence assumptions among $\w$ and determine the complexity of the normalizer $Z = \sum_{\w'}\exp S(\w')$. Fortunately, for chain-like interactions such as \eqref{eq:energy-function}, efficient dynamic programming routines are available. 

Two properties set our model apart from feedback-driven autoregressive models.  First, although $\psi$ captures only pairwise interactions, a state $\h_t$ will not only affect future observations but also all \emph{past} observations through the global coupling. Second, our model implicitly considers \emph{all} possible sequences $\w$ also at training time due to the global normalizer $Z$.

\subsection{Sampling}
\label{sec:sample}
Given a trained model, we can perform ancestral sampling via $\h\sim p(\h)$ and $\w\sim p(\w|\h)$. However, CRFs are undirected graphical models not designed with generation in mind and therefore we first need to derive ancestral sampling for $p(\w|\h)$. We can always write
$p(\w|\h)=\prod_t p(w_t|w_{1:t-1},\h)$ and find the factors
\begin{align}
	&\!\!\!p(w_t|w_{1:t-1},\h)\!=\!e^{\psi(w_{t-1}, w_t)+\psi(w_t)}\frac{\beta_{t+1}(w_t)}{\beta_t(w_{t-1})}\!\!\!\label{eq:marginal}\\
&\!\!\!\text{where}\notag\\
&\!\!\!\beta_t(w_{t-1})=\sum_{w_t}e^{\psi(w_{t-1},w_t) + \psi(w_t)}\beta_{t+1}(w_t) \label{eq:beta}
\end{align}
with special cases $\beta_1(w_0)=1$ and $\beta_{T+1}(w_T)=Z$ are the backwards probabilities we anyway need to compute for \eqref{eq:global-observations}. Not surprisingly, multiplying \eqref{eq:marginal} for $t=\Ts$ lets all $\beta$ terms cancel except for $1/Z$ and we recover \eqref{eq:global-observations}. However, this form is more amendable to sampling\footnote{In fact, one can train on \eqref{eq:marginal} instead of \eqref{eq:global-observations}. However, in our experiments we found the latter global normalization to be much more stable numerically.} and reveals an interesting property of globally normalized models: While the chain rule always allows to write such models autoregressively, we must expect a factor -- here $\beta_{t+1}(w_t)$ -- that implicitly marginalizes out future observations to assess compatibility with a specific next word $w_t$. Tractability of this factor is key to obtain a tractable model and is traded for expressiveness. While locally normalized models are on one end of the spectrum, a globally normalized with fully-connected potentials $\psi(\h(w))$ is on the other end. Such models employ an RNN in \emph{each} potential to obtain an un-normalized score $\psi$ from states $\h$ and have been investigated in conditional generation where argmax-decoding rather than sampling is requried \cite{wisemanR16}. Figure \ref{fig:sketch} shows the dependencies of the two extremes with our model in the middle.

\subsection{Embedding-based Local Correlations}
Often pairwise potentials can be parametrized directly, i.e.\ as $\psi(w_i,w_j)=\A_{ij}$ for some parameter matrix  $\A\in\mathbb R^{V\times V}$. However, in our setting this is problematic for two reasons. First, $|V|^2$ parameters are impractical in terms of model size for most vocabularies. Second, computations involving $\A$ are central to the complexity of computing log-likelihood during training. Namely, to compute the normalizer $Z$, we need to compute all $\beta$ quantities in \eqref{eq:beta}. Identifying $\beta_t(w_{t-1})$ as a $|V|$-dimensional vector $\bbeta_t$, we can write the summation in \eqref{eq:beta} as a matrix-vector product
\begin{align}
	\bbeta_t = \T( \oo_t \odot  \bbeta_{t+1})\label{eq:betavector}
\end{align} 
where $\odot$ is an element-wise product, $\oo_t$ are the unary potentials $\psi(w_t)$ written as a vector and $\T=\exp\A$ element-wise. We observe, computing $Z$ naively requires $\mathcal O(|V|^2T)$ operations.

To overcome the above shortcomings, we propose to factorize $\T$ as 
\begin{align}
\T=\X^\top\S(\h_{t-1},\h_t)\Y\label{eq:potential-factorized}
\end{align}
into context-independent $d$-dimensional embeddings $\X,\Y\in\mathbb R^{d\times|V|}$ and a context-dependent $d\times d$ interaction matrix computed by a neural network $\S : \R^{d'}\times\R^{d'}\rightarrow\R^{d\times d}$. This reduces the memory requirement to $\mathcal O(d|V|)$ and compute time to $\mathcal O(d|V|T)$, which is comparable to computing standard soft-max logits. As an additional benefit we can initialize $\X$ and $\Y$ with pre-trained word-embeddings, a technique often reported to improve convergence. Sine $\A$ does not have more structure than being strictly positive element-wise, it is sufficient to use strictly positive activation functions around the layers in \eqref{eq:betavector} to obtain a valid factorization.

\subsection{Training}
\label{sec:train}
As is standard for latent sequential models, we use variational inference for training~\cite{blei2017variational,zhang2018advances}. We introduce a parametrized approximate inference model $q(\h|\w)$ to maximize the evidence lower bound (ELBO) for a sampled trajectory instead of maximizing the marginal across all trajectories:
\begin{align}
	\log p(\w) &=\int p(\w,\h)d\h \\
	&\geq \E_{q}\left[ \log p(\w|\h) + \log \frac{p(\h)}{q(\h|\w)} \right]\label{eq:elbo}
\end{align}
The first term of \eqref{eq:elbo} measures reconstruction while the second measures the discrepancy between the trajectories implied by the inference model $q$ and the generative model $p$. The exact form of $p(\h)$ depends on its factorization and if it is autoregressive but for us simply $p(\h)=\prod_t p(\h_t|\h_{t-1})$, which casts us as an autoregressive extension of \citet{schmidt18}.
\paragraph{Inference model}
Like \cite{fraccaroSPW2016}, we choose $q$ to factorize as the true posterior
\begin{align}
	q(\h|\w)=\prod_{t=1}^T q(\h_t|\h_{t-1},w_{t:T})
\end{align}
where $w_{1:T}$ is encoded using an RNN running backwards in time to parameterize mean and variance of a Gaussian for $q(\h_t|\h_{t-1},w_{t:T})$.
%Together with $\h_{t-1}$, a feed-forward network then computes the parametrization, i.e.\ mean and variance, of $q(\h_t|\h_{t-1},w_{t:T})$.
For optimization we follow existing work \cite{fraccaroSPW2016, goyalSCKB17} and use the re-parametrization trick \cite{rezendeMW14, kingmaSW16} to perform a stochastic gradient step on \eqref{eq:elbo} with Adam \cite{kingmaB14} using a single trajectory.

\section{Experiments}
Exposure-bias can be summarized as over-confident conditioning on ``pseudo" predictions during training. The strength of the bias depends on the informativeness of such predictions, which in turn depends on the remaining context provided. 	

We test our proposed method on \emph{unconditional} generation which does not provide context such as a source sentence to narrow down possible outputs a priori. Hence, potential biases are more pronounced and generation is isolated from effects induced by i.e.\ a translation or summarization task.

\paragraph{Setup}
Unconditional generation is still considered a challenging task for both, GANs and latent stochastic models, \cite{fedusGD2018} and standard RNNs  form a very competitive baseline \cite{semeniuta18}. To obtain a homogeneous text dataset of low complexity we extract the plain text (text and hypothesis) from the Standard SNLI dataset \cite{bowman15snli} (For details and samples see Appendix \ref{appendix:data}).

\paragraph{Baselines}

We compare against a GRU (an LSTM performed on par) standard RNN of matching state size denoted  \textsc{dRNN}. We also include SeqGAN\footnote{We use the hyper-parameters recommended by the authors even though the state size is larger than ours.} \cite{yuZWY16}, a popular GAN architecture for unconditional generation. Further, we restrict our model to unary potentials to obtain a non-autoregressive state space model similar to that of \citet{schmidt18}, denoted \textsc{SSM}. Finally, \textsc{2-GRAM} is a bi-gram language model and \textsc{ORACLE} is held-out data, which represents the gold-standard for unconditional generation.

\paragraph{Parameterization}
We use 16-dimensional latent states, pre-train 100-dimensional GloVe embeddings and use word and context vectors for $\Y$ and $\X$. For $\S$ we found a diagonal matrix to perform best. In this case, the symmetry of $\T$  is broken by larger unary potentials.  While we find larger word embedding dimensionality to improve performance, the model does not benefit from more latent dimensions as an RNN does from hidden dimensions, a known issue of deep latent variable models \cite{schmidt18, ziegler19}.

\subsection{Qualitative Results}
Table \ref{table:cherry} shows selected output generated by our model (See Appendix \ref{appendix:samples} for more output). 
\begin{table}[h]
\centering
\def\arraystretch{1.}
\begin{tabular}{l}
\textit{a dog runs .}\\
\textit{the children are alone .}\\
\textit{the man is being beaten .}\\
\textit{the man is inside working onstage .}\\
\textit{the dog is outside with his girlfriend .}\\
\textit{two dogs going swimming in an open-air festival .}\\
\textit{a young lady wearing a pink shirt is studying .}\\
\end{tabular}
\caption{Output of our model of different length.}
\label{table:cherry}
\end{table}
While many of our sentences are grammatical and mimic those of the dataset we note that the corpus is not large enough to learn common sense and all models including the baselines sometimes generate output such as \textit{two men are burning snow}.

\subsection{Quantitative Results}
Perplexity under external language models is the standard metric to evaluate unconditional output \cite{fedusGD2018} and  we use Kneser-Ney-smoothed models up to\footnote{We find that the data is too sparse to train 4-gram language models as measured on a test-set.} $n=3$ estimated on the training data using SRILM \cite{srilm}.

In addition, we propose to estimate some important aggregate statistics easily verifiable against the real data. We choose length $l$ and percentage of unique sentences $\rho_\textsc{\tiny UNI}$ to assess diversity and percentage of token repetitions $\rho_\textsc{\tiny REP}$ to adress a failure mode often found in generative models \cite{tu16}. Table \ref{table:perplexity} shows the results.

\begin{table}[ht]
\centering
\small
\begin{tabular}{llllll}
&$\textsc{PPL}_2$ & $\textsc{PPL}_3$ & $\rho_\textsc{\tiny REP}$ & $l$ & $\rho_\textsc{\tiny UNI}$\\
\hline\hline
\textsc{\textbf{SSM+CRF}} & 40.1 & 41.9 &  0.35 & 8.4 & 98\\
\hline
\textsc{SSM} & 158.5& 172.2 & 9.20 & 8.9 & 100\\
\textsc{dRNN} & 47.1 & 43.5 & 0.78 & 8.7 & 99\\
\textsc{SeqGAN-20e} & 22.4 & 23.1 & 0.63 & 5.7 & 58\\
\textsc{SeqGAN-200e} & 53.0 & 57.0 & 6.88 & 7.1 & 80\\
\hline\hline
\textsc{2-GRAM} & 34.4 & 46.3 & 0.27 & 8.0 & 82\\
\textsc{ORACLE} & 26.7 & 17.7 & 0.17 & 8.9 & 99\\
\end{tabular}
\caption{Our model \textbf{SSM+CRF} evaluated against the baselines on 100K generated sentences each: Perplexity of  output under external language model $\textsc{PPL}_n$, percentage of repeated tokens per sentence $\rho_\textsc{\tiny REP}$, length $l$, and percentage of unique sentences $\rho_\textsc{\tiny UNI}$. All statistics should be compared to \textsc{ORACLE}, a held-out data split.}
\label{table:perplexity}
\end{table}

\section{Discussion and Future Work}
In terms of perplexity our model clearly improves over \textsc{SSM}, outperforms \textsc{dRNN} as measured by bigram statistics, and is on par with it in terms of trigram statistics. Of course, \textsc{2-GRAM} excels in terms of bigram statistics, yet falls behind on longer statistics. This confirms that our model can learn beyond pairwise interactions through the latent chain.
In addition, through our explicit model of pairwise interaction we obtain repetitions $\rho_\textsc{\tiny REP}$ significantly closer to the real data distribution.

For \textsc{SeqGAN} we report after 20 epochs (as used by the authors) and 200 epochs. We observe in general shorter output with more repetition (i.e.\ of words \textit{are}, \textit{is} and \textit{up}) and note that depending on training time the stellar fluency is traded with a significant bias on length $l$ and very poor diversity $\rho_\textsc{\tiny UNI}$, a tendency also observed by \citet{xu2018} and possibly related to the choice of temperature parameter \cite{Caccia18}. While it is not our goal to provide a deeper analysis of GANs here, the example shows how unconditional generation can reveal tradeoffs not present in a conditional setting.

\paragraph{Future Work}
We have shown that autoregressive predictions expressed in the observation model instead of hidden states deliver better results on a simple corpus. In particular, mistakes at the bigram-level, such as repetitions, are avoided and we suspect that more densely connected CRFs allow to extend these promising results to more complex patterns found in more complex corpora. In future work we plan to investigate if CRF variants such as \cite{belangerYM17} or  \cite{kraehenbuehl12} can be adapted to allow efficient sampling and to scale to word vocabulary sizes.

\section{Conclusion}

We have shown an alternative methodology to autoregressive modeling that avoids exposure-bias in hidden states by design through a globally normalized observation model. We derived a sampling method and an efficient embedding-based parameteriation of CRFs to trade expressiveness with tractability. On an unconditional generation task, we obtain better results than a deterministic RNN in a low-dimensional setting and more consistent results than a GAN baseline. Finally, we have pointed into directions on how to capture more complex correlations.

\newpage

\bibliographystyle{acl_natbib}
\bibliography{bibliography}

\appendix

\onecolumn

\section{Data}
\label{appendix:data}

We use a 15K vocabulary (\texttt{unk}-rate 0.5\%) and limit the length to 15 tokens (9\% sentences are longer) resulting  in a 540K sentences (4.8M tokens) dataset which is about six times the size of the Penn Treebank corpus. An excerpt of the (\texttt{unk}ed) training data is shown below.

\newlength{\mylen}
\settowidth{\mylen}{\textsc{dRNN-64d}}

\paragraph{\textsc{train-set}}
\begin{tabular}{l}
\textit{a child is sitting inside and watching tv .}\\
\textit{the man is wearing worn out jeans .}\\
\textit{two motorcyclists are in physical contact .}\\
\textit{a girl colors with markers .}\\
\textit{nighttime scene of outdoor food kiosk .}\\
\textit{man playing trumpet on the street for dollar bills .}\\
\textit{there is a sign at the market .}\\
\textit{people relaxing on a beach .}\\
\textit{the girl is riding in the back seat of the car .}\\
\textit{the choir is at the bowling alley .}\\
\textit{a man is naked outside .}\\
\textit{the woman is interested in art .}\\
\textit{a man is cooking .}\\
\textit{three asian men conversing while sitting against a wall .}\\
\textit{a cat \_unk at a sheep .}\\
\textit{child shovels snow , near bush , outside of building .}\\
\textit{a man holds a bird by the train .}\\
\textit{a woman sits at home looking at porn .}\\
\textit{women playing a card game .}\\
\textit{children are eating apples at their school .}\\
\textit{three women wait by a cart of axes at a carnival .}\\
\textit{a man with a pierced eyebrow is looking at a laptop screen , while a cat is in the background .}\\
\textit{the asian man is working hard on some documents .}\\
\textit{people fill balloons .}\\
\textit{a man stands next to a bank of computer gambling machines .}\\
\textit{a volkswagon bug is driving in a field .}\\
\textit{people barely visible except for upraised arms .}\\
\textit{young men enjoy the shade under the deck while at the beach .}\\
\textit{a musician performs inside .}\\
\textit{5 people are in an airplane .}\\
\textit{the dog chases a squirrel up a tree .}\\
\textit{tall humans performing surgery .}\\
\textit{a boy and a girl are playing with dolls in their backyard .}\\
\textit{several people play basketball outside .}\\
\textit{a man looks at the camera while he sits down in a cluttered house .}\\
\textit{two boys skateboard in front of a large statue of a soldier riding a horse in a town plaza .}\\
\textit{the men are asleep .}\\
\textit{the couple are waiting at the movies .}\\
\end{tabular}

\section{Example Output}
\label{appendix:samples}

We report 20 unfiltered generated sentences for each of our proposed models as well as the RNN baselines. For a more representative analysis, please refer to \url{https://github.com/schmiflo/crf-generation} where we provide 100K sentences generated by the models along with the reference SRI $n$-gram models. 

\begin{table}[hb]
\paragraph{\textsc{dRNN-16d}}
\begin{tabular}{l}
\textit{a group of people are directing vehicles .}\\
\textit{a portrait of men .}\\
\textit{men walk horses .}\\
\textit{a dog runs over a towel slide down the hill .}\\
\textit{three women waiting for the bus .}\\
\textit{a farmer takes a break .}\\
\textit{two white dogs are rolling on the grass .}\\
\textit{a clown is wearing mittens .}\\
\textit{the man talks away .}\\
\textit{two women are throwing yogurt .}\\
\textit{young pick camels with it face in the sand .}\\
\textit{a classroom of people vendors tacos bowling and protective device .}\\
\textit{the divers are being done .}\\
\textit{a girl is spinning .}\\
\textit{a person is wearing a dark white sling mower .}\\
\textit{the woman and a woman are displaying leisure together .}\\
\textit{a girl jumping to the skate park .}\\
\textit{young firefighters are getting a cookout up .}\\
\textit{man looking horse on the river .}\\
\textit{more they went competing .}\\
\end{tabular}
\end{table}

\begin{table}[hb]
\paragraph{\textsc{SSM\phantom{blaa}}}
\begin{tabular}{l}
\textit{the man sits the a t-shirt and her snacks from traditional package through work .}\\
\textit{some are all music ride behind his the pool .}\\
\textit{people smiling around each one a celebrating down their trail .}\\
\textit{someone one knives at the cellphone camping video living scaffold .}\\
\textit{two waiting to white lawyer in a locker organized open .}\\
\textit{a man is performing under food .}\\
\textit{he was sitting at home other .}\\
\textit{several scientist path in snowboard of new hands at to .}\\
\textit{people are a doorbell .}\\
\textit{people standing performing down to other .}\\
\textit{it employees looking a groom are taking front exhibit}\\
\textit{homeless man talking down men clouds at their point .}\\
\textit{a child rides the some equipment .}\\
\textit{men doing ice pink ball .}\\
\textit{the ball is close \_unk . .}\\
\textit{the indian is wishing and and out as use large ice of fashion '' .}\\
\textit{puppies riding a movie is weaving the grass driveway}\\
\textit{two women pose to mexican new reception is playing}\\
\textit{there of man are watching the park .}\\
\textit{young are singing with hard rowing among couch on}\\

\end{tabular}
\end{table}

\begin{table}[hb]
\paragraph{\textsc{SSM+CRF}}
\begin{tabular}{l}
\textit{the band plays checkers on the beach .}\\
\textit{an artist is swimming .}\\
\textit{a football player number plaid shirt together .}\\
\textit{a cat is in a gym park nearby .}\\
\textit{a child asked does a frozen pole .}\\
\textit{cute greyhound train .}\\
\textit{a group of people in his newspaper .}\\
\textit{a human is selling fun .}\\
\textit{nobody is walking past each other on a bowl .}\\
\textit{a man takes time looking water with a helmet is hungry .}\\
\textit{a pig is near his face .}\\
\textit{a person is holding dark .}\\
\textit{a man and gold paint a wedding condition .}\\
\textit{two people are dancing in strike a bookcase .}\\
\textit{there is a gray-haired asian man pose for some art .}\\
\textit{a person standing on a roller coaster .}\\
\textit{the lighthouse .}\\
\textit{a couple gathered for his dog outside .}\\
\textit{a group choir dance .}\\
\textit{people building is standing next to her family part of a gun .}\\
\end{tabular}
\end{table}

\begin{table}[hb]
\paragraph{\textsc{SeqGAN-20e}}
\begin{tabular}{l}
\textit{two people are attend next .}\\
\textit{a man is running in dark .}\\
\textit{many people are on their at metal .}\\
\textit{there are women on .}\\
\textit{two boys are in green .}\\
\textit{they are at their like kids .}\\
\textit{filled person is in their large with items .}\\
\textit{two indian men are chasing .}\\
\textit{three people are taking pictures outside .}\\
\textit{the boy is parked .}\\
\textit{a boy is walking to the wedding .}\\
\textit{the man is shopping .}\\
\textit{a man is \_unk .}\\
\textit{a boy in a girl .}\\
\textit{a boy is on their next in the mountains .}\\
\textit{the man is drinking .}\\
\textit{one man is to sitting .}\\
\textit{a man is performing a yoga .}\\
\textit{a couple is swimming in the camera .}\\
\textit{the woman is .}\\
\end{tabular}
\end{table}

\begin{table}[hb]
\paragraph{\textsc{SeqGAN-200e}}
\begin{tabular}{l}
\textit{the girls are a blonde talk .}\\
\textit{a workers are very portraits .}\\
\textit{football are up in yellow .}\\
\textit{child in vegetables .}\\
\textit{a girl is in costumes .}\\
\textit{children are waiting for something .}\\
\textit{woman are eating picture .}\\
\textit{three brothers are performing for eyebrows .}\\
\textit{dirt man is are in number apples .}\\
\textit{a boy is in son .}\\
\textit{i eat nice hugging .}\\
\textit{the runners are school .}\\
\textit{the couple enjoy their dodgeball .}\\
\textit{two guard are their bicycle .}\\
\textit{individuals are children at their bmx .}\\
\textit{dog chasing some kicking .}\\
\textit{outer are are around on .}\\
\textit{the brother is wearing seated are people with outside .}\\
\textit{bride are up .}\\
\textit{children are are taking pictures of gray .}\\
\end{tabular}
\end{table}

\end{document}